\documentclass[11pt,a4paper]{article}
\usepackage[hyperref]{acl2020}
\usepackage{times}
\usepackage{latexsym}
\usepackage{amsmath,amssymb,amsfonts,bm}
\usepackage{multirow}
\usepackage{booktabs}
\usepackage{graphicx}
\usepackage[margin=5pt]{subcaption}

\DeclareMathOperator*{\argmax}{arg\,max}

\usepackage{microtype}

\aclfinalcopy 
\def\aclpaperid{1220} 

\title{Instance-Based Learning of Span Representations:\\A Case Study through Named Entity Recognition}

\author{Hiroki Ouchi$^{1,2}$ \hspace{0.7cm} Jun Suzuki$^{2,1}$  \hspace{0.7cm} Sosuke Kobayashi$^{2,3}$ \\
\textbf{Sho Yokoi}$^{2,1}$ \hspace{0.7cm} \textbf{Tatsuki Kuribayashi}$^{2,4}$  \hspace{0.7cm} \textbf{Ryuto Konno}$^2$  \hspace{0.7cm}\textbf{Kentaro Inui}$^{2,1}$\\
  {$^1$ RIKEN} \hspace{0.7cm} {$^2$ Tohoku University} \hspace{0.7cm} {$^3$ Preferred Networks, Inc.} \hspace{0.7cm} {$^4$ Langsmith, Inc.}\\
  {\tt hiroki.ouchi@riken.jp} \\ \{{\tt jun.suzuki,sosk,yokoi,kuribayashi,ryuto,inui\}@ecei.tohoku.ac.jp}}

\date{}

\begin{document}
\maketitle
\begin{abstract}
Interpretable rationales for model predictions play a critical role in practical applications.
In this study, we develop models possessing interpretable inference process for structured prediction.
Specifically, we present a method of instance-based learning that learns similarities between spans.
At inference time, each span is assigned a class label based on its similar spans in the training set, where it is easy to understand how much each training instance contributes to the predictions.
Through empirical analysis on named entity recognition, we demonstrate that our method enables to build models that have high interpretability without sacrificing performance.
\end{abstract}

\section{Introduction}
\label{sec:intro}

Neural networks have contributed to performance improvements in structured prediction.
Instead, the rationales underlying the model predictions are difficult for humans to understand~\cite{lei-etal-2016-rationalizing}.
In practical applications,  interpretable rationales play a critical role for driving human's decisions and promoting human-machine cooperation~\cite{ribeiro2016should}.
With this motivation, we aim to build models that have high interpretability without sacrificing performance.
As an approach to this challenge, we focus on {\it instance-based learning}.

Instance-based learning~\cite{aha1991instance} is a machine learning method that learns similarities between instances.
At inference time, the class labels of the most similar training instances are assigned to the new instances.
This transparent inference process provides an answer to the following question: \textit{Which points in the training set most closely resemble a test point or influenced the prediction?}
This is categorized into \textit{example-based explanations}~\cite{plumb2018model,baehrens2010explain}.
Recently, despite its preferable property, it has received little attention and been underexplored.

This study presents and investigates an instance-based learning method for \textit{span representations}.
A span is a unit that consists of one or more linguistically linked words.
\textit{Why do we focus on spans instead of tokens?}
One reason is relevant to performance.
Recent neural networks can induce good span feature representations and achieve high performance in structured prediction tasks, such as named entity recognition (NER)~\cite{sohrab-miwa-2018-deep,xia-etal-2019-multi}, constituency parsing~\cite{stern-etal-2017-minimal,kitaev-etal-2019-multilingual}, semantic role labeling (SRL)~\cite{he-etal-2018-jointly,ouchi-etal-2018-span} and coreference resolution~\cite{lee-etal-2017-end}.
Another reason is relevant to interpretability.
The tasks above require recognition of linguistic structure that consists of spans.
Thus, directly classifying each span based on its representation is more interpretable than token-wise classification such as BIO tagging, which reconstructs each span label from the predicted token-wise BIO tags.

Our method builds a feature space where spans with the same class label are close to each other.
At inference time, each span is assigned a class label based on its neighbor spans in the feature space.
We can easily understand why the model assigned the label to the span by looking at its neighbors.
Through quantitative and qualitative analysis on NER, we demonstrate that our instance-based method enables to build models that have high interpretability and performance.
To sum up, our main contributions are as follows.
\vspace{-0.4cm}
\begin{itemize}
\setlength{\parskip}{0cm} 
\setlength{\itemsep}{0.1cm} 
\item This is the first work to investigate instance-based learning of span representations.\footnote{Our code is publicly available at \url{https://github.com/hiroki13/instance-based-ner.git}.}
\item Through empirical analysis on NER, we demonstrate our instance-based method enables to build models that have high interpretability without sacrificing performance.
\end{itemize}

\section{Related Work}
\label{sec:rwork}

Neural models generally have a common technical challenge: the black-box property.
The rationales underlying the model predictions are opaque for humans to understand.
Many recent studies have tried to look into classifier-based neural models \cite{ribeiro2016should,lundberg2017unified,koh2017understanding}.
In this paper, instead of looking into the black-box, we build interpretable models based on instance-based learning.

Before the current neural era, instance-based learning, sometimes called memory-based learning \cite{daelemans2005memory}, was widely used for various NLP tasks, such as part-of-speech tagging \cite{daelemans-etal-1996-mbt}, dependency parsing \cite{nivre-etal-2004-memory} and machine translation \cite{nagao1984framework}.
For NER, some instance-based models have been proposed \cite{tjong-kim-sang-2002-memory,de-meulder-daelemans-2003-memory,hendrickx-van-den-bosch-2003-memory}.
Recently, despite its high interpretability, this direction has not been explored.

One exception is \citet{wiseman-stratos-2019-label}, which used instance-based learning of token representations.
Due to BIO tagging, it faces one technical challenge: inconsistent label prediction.
For example, an entity candidate ``World Health Organization" can be assigned inconsistent labels such as ``{\tt B-LOC} {\tt I-ORG} {\tt I-ORG}," whereas the ground-truth labels are ``{\tt B-ORG} {\tt I-ORG} {\tt I-ORG}."
To remedy this issue, they presented a heuristic technique for encouraging contiguous token alignment.
In contrast to such token-wise prediction, we adopt span-wise prediction, which can naturally avoid this issue because each span is assigned one label.

NER is generally solved as (i) sequence labeling or (ii) span classification.\footnote{Very recently, a hybrid model of these two approaches has been proposed by \citet{liu-etal-2019-towards}.}
In the first approach, token features are induced by using neural networks and fed into a classifier, such as conditional random fields \cite{lample-etal-2016-neural,ma-hovy-2016-end,chiu-nichols-2016-named}.
One drawback of this approach is the difficulty dealing with nested entities.\footnote{Some studies have sophisticated sequence labeling models for nested NER~\cite{ju-etal-2018-neural,zheng-etal-2019-boundary}.}
By contrast, the span classification approach, adopted in this study, can straightforwardly solve nested NER \cite{finkel-manning-2009-nested,sohrab-miwa-2018-deep,xia-etal-2019-multi}.\footnote{There is an approach specialized for nested NER using hypergraphs \cite{lu-roth-2015-joint,muis-lu-2017-labeling,katiyar-cardie-2018-nested,wang-lu-2018-neural}.}

\section{Instance-Based Span Classification}
\label{sec:method}

\subsection{NER as span classification}
\label{sec:span-class}
NER can be solved as multi-class classification, where each of possible spans in a sentence is assigned a class label.
As we mentioned in Section~\ref{sec:rwork}, this approach can naturally avoid inconsistent label prediction and straightforwardly deal with nested entities.
Because of these advantages over token-wise classification, span classification has been gaining a considerable attention~\cite{sohrab-miwa-2018-deep,xia-etal-2019-multi}.

Formally, given an input sentence of $T$ words $X = (w_1, w_2, \dots, w^{}_T)$, we first enumerate possible spans $\mathcal{S}(X)$, and then assign a class label $y \in \mathcal{Y}$ to each span $s \in \mathcal{S}(X)$.
We will write each span as $s = (a, b)$, where $a$ and $b$ are word indices in the sentence: $1 \le a \le b \le T$.
Consider the following sentence.\\

\vspace{-0.3cm}
\noindent
\hspace{0.8cm} Franz$_1$ \hspace{0.2cm} Kafka$_2$ \hspace{0.2cm} is$_3$ \hspace{0.2cm} a$_4$ \hspace{0.2cm} novelist$_5$

\noindent
\hspace{0.8cm} [\hspace{0.65cm} {\tt PER} \hspace{0.65cm}] \\

\vspace{-0.3cm}
\noindent
Here, the possible spans in this sentence are $\mathcal{S}(X) = \{ (1,1), (1,2), (1, 3), \dots, (4,5), (5,5) \}$.
``Franz~Kafka," $s = (1, 2)$, is assigned the person type entity label ($y = \texttt{PER}$).
Note that the other non-entity spans are assigned the null label ($y = \texttt{NULL}$).
For example, ``a novelist," $s = (4, 5)$, is assigned \texttt{NULL}.
In this way, the \texttt{NULL} label is assigned to non-entity spans, which is the same as the {\tt O} tag in the BIO tag set.

The probability that each span $s$ is assigned a class label $y$ is modeled by using softmax function:
\begin{align}
\nonumber
\text{P}(y | s) & = \frac{\text{exp}(\text{score}(s, y))}{\displaystyle \sum_{y' \in \mathcal{Y}} \text{exp}(\text{score}(s, y'))} \:\:.
\end{align}

\noindent
Typically, as the scoring function, the inner product between each label weight vector ${\bf w}_y$ and span feature vector ${\bf h}_s$ is used:
\begin{align}
\nonumber
\text{score}(s, y) = {\bf w}_y \cdot {\bf h}_s \:\:.
\end{align}
\noindent
The score for the \texttt{NULL} label is set to a constant, $\text{score}(s, y=\texttt{NULL}) = 0$, similar to logistic regression~\cite{he-etal-2018-jointly}.
For training, the loss function we minimize is the negative log-likelihood:
\begin{equation}
\nonumber
\mathcal{L} = - \sum_{(X, Y) \in \mathcal{D}} \sum_{(s, y) \in \mathcal{S}(X, Y)} \text{log} \: \text{P}(y | s) \:\:,
\end{equation}

\noindent
where $\mathcal{S}(X, Y)$ is a set of pairs of a span $s$ and its ground-truth label $y$.
We call this kind of models that use label weight vectors for classification {\it classifier-based span model}.

\begin{figure*}[t]
  \begin{center}
    \includegraphics[width=15cm]{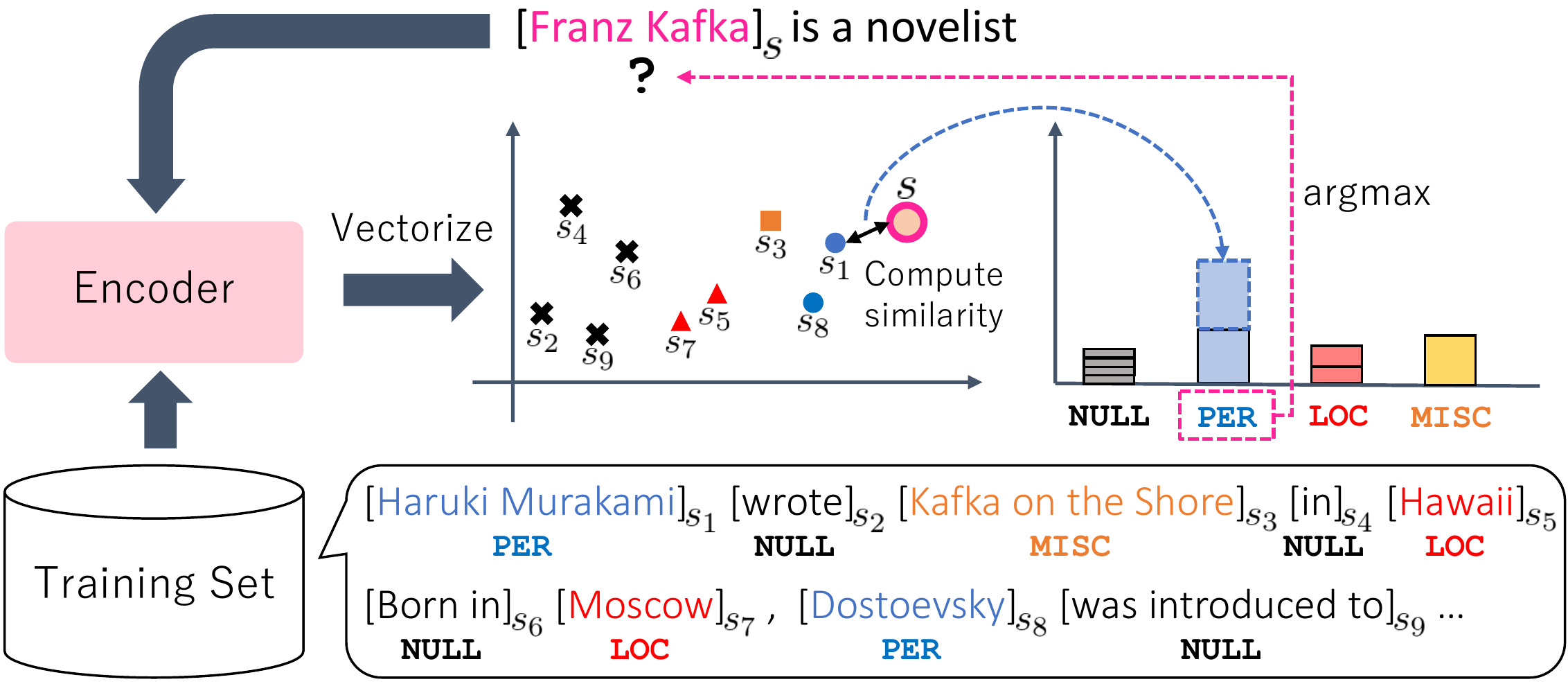}
  \end{center}
  \caption{\label{fig:example} Illustration of our instance-based span model. An entity candidate ``Franz Kafka" is used as a query and vectorized by an encoder. In the vector space, similarities between all pairs of the candidate ($s$) and the training instances ($s_1, s_2, \dots, s_9$) are computed, respectively. Based on the similarities, the label probability (distribution) is computed, and the label with the highest probability {\tt PER} is assigned to ``Franz Kafka."}
\end{figure*}

\subsection{Instance-based span model}
\label{sec:model}

Our \textit{instance-based span model} classifies each span based on similarities between spans.
In Figure~\ref{fig:example}, an entity candidate ``Franz~Kafka" and the spans in the training set are mapped onto the feature vector space, and the label distribution is computed from the similarities between them.
In this inference process, it is easy to understand how much each training instance contributes to the predictions.
This property allows us to explain the predictions by specific training instances, which is categorized into \textit{example-based explanations} \cite{plumb2018model}.

Formally, within the neighbourhood component analysis framework \cite{goldberger2005neighbourhood}, we define the {\it neighbor span probability} that each span $s_i \in \mathcal{S}(X)$ will select another span $s_j$ as its neighbor from candidate spans in the training set:

\begin{align}
\label{eq:p}
\text{P}(s_j | s_i, \mathcal{D'}) = \frac{\text{exp}(\text{score}(s_i, s_j))}{\displaystyle \sum_{s_k \in \mathcal{S}(D')} \text{exp}(\text{score}(s_i, s_k))} \:\:.
\end{align}

\noindent
Here, we exclude the input sentence $X$ and its ground-truth labels $Y$ from the training set $\mathcal{D}$: $\mathcal{D}' = \mathcal{D} \setminus \{(X,Y)\}$, and regard all other spans as candidates: $\mathcal{S}(\mathcal{D}') = \{ s \in \mathcal{S}(X') \: | \: (X', Y') \in \mathcal{D}' \}$.
The scoring function returns a similarity between the spans $s_i$ and $s_j$.
Then we compute the probability that a span $s_i$ will be assigned a label $y_i$:
\begin{equation}
\label{eq:cond-proba}
\text{P}(y_i | s_i) =\displaystyle \sum_{s_j \in \mathcal{S}(\mathcal{D}', y_i)} \text{P}(s_j | s_i, \mathcal{D}') \:\:.
\end{equation}

\noindent
Here, $\mathcal{S}(\mathcal{D}', y_i) = \{ s_j \in \mathcal{D}' | \: y_i = y_j \}$, so the equation indicates that we sum up the probabilities of the neighbor spans that have the same label as the span $s_i$.
The loss function we minimize is the negative log-likelihood:
\begin{equation}
\nonumber
\mathcal{L} = - \sum_{(X, Y) \in \mathcal{D}} \sum_{(s_i, y_i) \in \mathcal{S}(X, Y)} \text{log} \: \text{P}(y_i | s_i) \:\:,
\end{equation}

\noindent
where $\mathcal{S}(X, Y)$ is a set of pairs of a span $s_i$ and its ground-truth label $y_i$.
At inference time, we predict $\hat{y}_i$ to be the class label with maximal marginal probability:
\begin{equation}
\nonumber
\hat{y}_i = \argmax_{y \in \mathcal{Y}} \text{P}(y | s_i)  \:\:,
\end{equation}

\noindent
where the probability $\text{P}(y | s_i)$ is computed for each of the label set $y \in \mathcal{Y}$.

\paragraph{Efficient neighbor probability computation}
The neighbor span probability $\text{P}(s_j | s_i, \mathcal{D}')$ in Equation~\ref{eq:p} depends on the entire training set $\mathcal{D}'$, which leads to heavy computational cost.
As a remedy, we use random sampling to retrieve $K$ sentences $\mathcal{D}'' = \{ (X'_k, Y'_k) \}^K_{k=0}$ from the training set $\mathcal{D}'$.
At training time, we randomly sample $K$ sentences for each mini-batch at each epoch.
This simple technique realizes time and memory efficient training.
In our experiments, it takes less than one day to train a model on a single GPU\footnote{NVIDIA DGX-1 with Tesla V100.}.

\section{Experiments}
\label{sec:exp}

\subsection{Experimental setup}
\paragraph{Data}
We evaluate the span models through two types of NER: (i) flat NER on the CoNLL-2003 dataset~\cite{tjong-kim-sang-de-meulder-2003-introduction} and (ii) nested NER on the GENIA dataset\footnote{We use the same one pre-processed by \citet{zheng-etal-2019-boundary} at \url{https://github.com/thecharm/boundary-aware-nested-ner}}~\cite{kim2003genia}.
We follow the standard training-development-test splits.

\paragraph{Baseline}
We use a classifier-based span model (Section~\ref{sec:span-class}) as a baseline.
Only the difference between the instance-based and classifier-based span models is whether to use softmax classifier or not.

\paragraph{Encoder and span representation}
We adopt the encoder architecture proposed by \citet{ma-hovy-2016-end}, which encodes each token of the input sentence $w_t \in X$ with word embedding and character-level CNN.
The encoded token representations ${\bf w}_{1:T} = (\textbf{w}_1, \textbf{w}_2, \dots, \textbf{w}_T)$ are fed to bidirectional LSTM for computing contextual ones $\overrightarrow{\bf h}_{1:T}$ and $\overleftarrow{\bf h}_{1:T}$.
From them, we create ${\bf h}^\text{lstm}_s$ for each span $s = (a, b)$ based on LSTM-minus~\cite{wang-chang-2016-graph}.
For flat NER, we use the representation $\mathbf{h}^\mathrm{lstm}_s =  [\overrightarrow{\mathbf{h}}_b- \overrightarrow{\mathbf{h}}_{a-1}, \overleftarrow{\mathbf{h}}_a - \overleftarrow{\mathbf{h}}_{b+1}]$.
For nested NER, we use $\mathbf{h}^\mathrm{lstm}_s =  [\overrightarrow{\mathbf{h}}_b- \overrightarrow{\mathbf{h}}_{a-1}, \overleftarrow{\mathbf{h}}_a - \overleftarrow{\mathbf{h}}_{b+1}, \overrightarrow{\mathbf{h}}_a+ \overrightarrow{\mathbf{h}}_b, \overleftarrow{\mathbf{h}}_a+ \overleftarrow{\mathbf{h}}_b]$.\footnote{We use the different span representation from the one used for flat NER because concatenating the addition features, $\overrightarrow{\mathbf{h}}_a+ \overrightarrow{\mathbf{h}}_b$ and $\overleftarrow{\mathbf{h}}_a+ \overleftarrow{\mathbf{h}}_b$, to the subtraction features improves performance in our preliminary experiments.}
We then multiply $\mathbf{h}^\mathrm{lstm}_s$ with a weight matrix $\mathbf{W}$ and obtain the span representation: $\mathbf{h}_s = \mathbf{W} \: \mathbf{h}^\mathrm{lstm}_s$.
For the scoring function in Equation~\ref{eq:p} in the instance-based span model, we use the inner product between a pair of span representations: $\text{score}(s_i, s_j) = {\bf h}_{s_i} \cdot {\bf h}_{s_j}$.

\paragraph{Model configuration}
We train instance-based models by using $K=50$ training sentences randomly retrieved for each mini-batch.
At test time, we use $K=50$ nearest training sentences for each sentence based on the cosine similarities between their sentence vectors\footnote{For each sentence $X = (w_1, w_2, \dots, w^{}_T)$, its sentence vector is defined as the vector averaged over the word embeddings (GloVe) within the sentence: $\frac{1}{T}\sum_t \textbf{w}^\mathrm{emb}_t$.}.
For the word embeddings, we use the GloVe 100-dimensional embeddings~\cite{pennington-etal-2014-glove} and the BERT embeddings~\cite{devlin-etal-2019-bert}.\footnote{Details on the experimental setup are described in Appendices~\ref{sec:model-config}.}

\subsection{Quantitative analysis}
\label{sec:quantitative-analysis}
\begin{table}[t]
  \centering
  {\small
  \begin{tabular}{lcc} \toprule
  		    & Classifier-based & Instance-based \\ \midrule
                  \multicolumn{3}{c}{GloVe} \\ \midrule
                  Flat NER & 90.68 {\scriptsize $\pm$0.25} & 90.73  {\scriptsize $\pm$0.07} \\ 
                  Nested NER & 73.76 {\scriptsize $\pm$0.35} & 74.20 {\scriptsize $\pm$0.16} \\ \midrule
                  \multicolumn{3}{c}{BERT} \\ \midrule
                  Flat NER & 90.48 {\scriptsize $\pm$0.18} & 90.48  {\scriptsize $\pm$0.07} \\ 
                  Nested NER & 73.27 {\scriptsize $\pm$0.19} & 73.92 {\scriptsize $\pm$0.20} \\ \bottomrule
  \end{tabular}
  }
  \caption{Comparison between classifier-based and instance-based span models. Cells show the F$_1$ scores and standard deviations on each test set.}
  \label{tab:main-result}
\end{table}

\begin{figure}[t]
  \begin{center}
    \includegraphics[width=7cm]{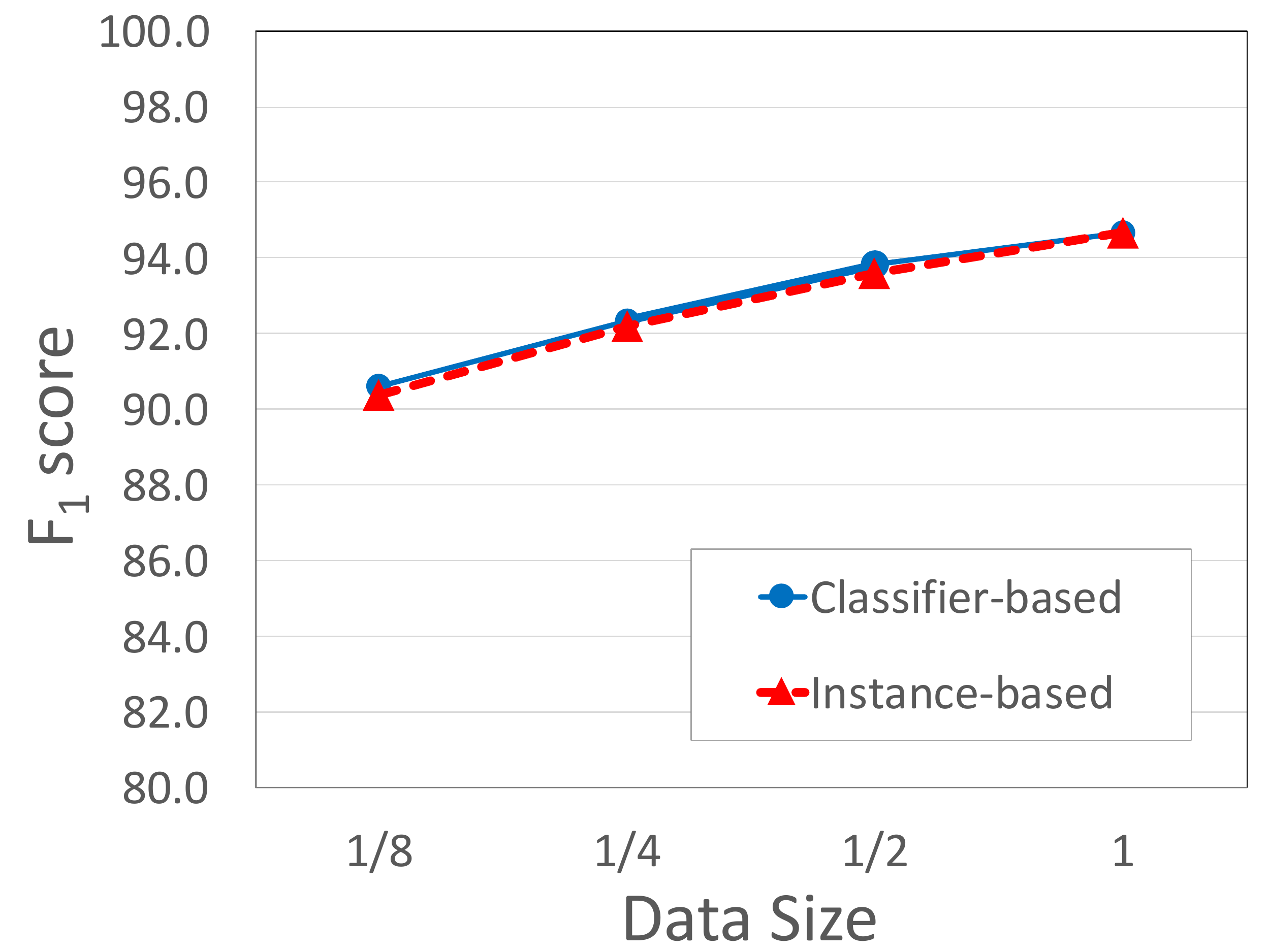}
  \end{center}
  \caption{Performance on the CoNLL-2003 development set for different amounts of the training set.}
  \label{fig:sampling}
\end{figure}

We report averaged F$_1$ scores across five different runs of the model training with random seeds.

\paragraph{Overall F$_1$ scores}
We investigate whether or not our instance-based span model can achieve competitive performance with the classifier-based span model.
Table~\ref{tab:main-result} shows F$_1$ scores on each test set.\footnote{The models using GloVe yielded slightly better results than those using BERT. One possible explanation is that subword segmentation is not so good for NER. In particular, tokens in upper case are segmented into too small elements, e.g., ``LEICESTERSHIRE" $\rightarrow$ ``L," ``\#\#EI," ``\#\#CE," ``\#\#ST," ``\#\#ER," ``\#\#S," ``\#\#H," ``\#\#IR," ``\#\#E."}
Consistently, the instance-based span model yielded comparable results to the classifier-based span model.
This indicates that our instance-based learning method enables to build NER models without sacrificing performance.

\paragraph{Effects of training data size}
Figure~\ref{fig:sampling} shows F$_1$ scores on the CoNLL-2003 development set by the models trained on full-size, $1/2$, $1/4$ and $1/8$ of the training set.
We found that (i) performance of both models gradually degrades when the size of the training set is smaller and (ii) both models yield very competitive performance curves.

\subsection{Qualitative analysis}
\label{sec:qualitative-analysis}

To better understand model behavior, we analyze the instance-based model using GloVe in detail.

\paragraph{Examples of retrieved spans}

\begin{table}[t]
  \centering
  {\footnotesize
  \begin{tabular}{rcl} \toprule
  \multicolumn{2}{r}{\textsc{Query}} &  ... \textbf{[}Tom Moody\textbf{]} took six for 82 but ... \\ \midrule
  \multicolumn{3}{c}{Classifier-based}\\ \midrule
1 & \texttt{PER} & ... \textbf{[}Billy Mayfair\textbf{]} and Paul Goydos and  ... \\
2 & \texttt{NULL} & ... \textbf{[}Billy Mayfair and Paul Goydos\textbf{]} and  ... \\
3 & \texttt{NULL} & ... \textbf{[}Billy Mayfair and Paul Goydos and\textbf{]}  ... \\
4 & \texttt{NULL} & ... \textbf{[}Billy\textbf{]} Mayfair and Paul Goydos and  ... \\
5 & \texttt{NULL} & ... \textbf{[}Ducati rider Troy Corser\textbf{]} , last year  ... \\ \midrule
  \multicolumn{3}{c}{Instance-based}\\ \midrule
1 & \texttt{PER} & \textbf{[}Ian Botham\textbf{]} began his test career ... \\
2 & \texttt{PER} & ... \textbf{[}Billy Mayfair\textbf{]} and Paul Goydos and  ... \\
3 & \texttt{PER} & ... \textbf{[}Mark Hutton\textbf{]} scattered four hits ... \\
4 & \texttt{PER} & ... \textbf{[}Steve Stricker\textbf{]} , who had a 68 , and  ...\\
3 & \texttt{PER} & ... \textbf{[}Darren Gough\textbf{]} polishing off ... \\ \bottomrule
  \end{tabular}
  }
  \caption{Example of span retrieval. An entity candidate ``Tom Moody" in the CoNLL-2003 development set used as a query for retrieving five nearest neighbors from the training set.}
  \label{tab:knn}
\end{table}

The span feature space learned by our method can be applied to various downstream tasks.
In particular, it can be used as a span retrieval system.
Table~\ref{tab:knn} shows five nearest neighbor spans of an entity candidate ``Tom Moody."
In the classifier-based span model, person-related but non-entity spans were retrieved.
By contrast, in the instance-based span model, person ({\tt  PER}) entities were consistently retrieved.\footnote{The query span ``Tom moody" was a cricketer at that time, and some neighbors, ``Ian Botham" and ``Darren Gough," were also cricketers.}
This tendency was observed in many other cases, and we confirmed that our method can build preferable feature spaces for applications.

\paragraph{Errors analysis}
\begin{table}[t]
  \centering
  {\footnotesize
  \begin{tabular}{rcl} \toprule
  \multicolumn{2}{r}{\textsc{Query}} & ... spokesman for \textbf{[}Air France\textbf{]} 's  ... \\
  \multicolumn{3}{c}{\hspace{2.2cm} Pred: \texttt{LOC}}\\
  \multicolumn{3}{c}{\hspace{2.2cm} Gold: \texttt{ORG}}\\ \midrule
1 & \texttt{LOC} & ... \textbf{[}Colombia\textbf{]} turned down American 's ... \\
2 & \texttt{LOC} & ... involving \textbf{[}Scotland\textbf{]} , Wales , ... \\
3 & \texttt{LOC} & ... signed in \textbf{[}Nigeria\textbf{]} 's capital Abuja ...\\
4 & \texttt{LOC} & ... in the West Bank and \textbf{[}Gaza\textbf{]} . \\
5 & \texttt{LOC} & ... on its way to \textbf{[}Romania\textbf{]}  ... \\ \bottomrule
  \end{tabular}
  }
  \caption{Example of an error by the instance-based span model. Although the gold label is \texttt{ORG} (Organization), the wrong label \texttt{LOC} (Location) is assigned. }
  \label{tab:error}
\end{table}

The instance-based span model tends to wrongly label spans that includes location or organization names.
For example, in Table~\ref{tab:error}, the wrong label \texttt{LOC} (Location) is assigned to ``Air France" whose gold label is \texttt{ORG} (Organization).
Note that by looking at the neighbors, we can understand that country or district entities confused the model.
This implies that prediction errors are easier to analyze because the neighbors are the rationales of the predictions.

\subsection{Discussion}
\label{sec:discussion}

\begin{table}[t]
  \centering
  {\small
  \begin{tabular}{lcc} \toprule
  		    & Classifier-based & Instance-based \\ \midrule
                  GloVe & 94.91 {\scriptsize $\pm$0.11} & 94.96  {\scriptsize $\pm$0.06} \\ 
                  BERT & 96.20 {\scriptsize $\pm$0.03} & 96.24  {\scriptsize $\pm$0.04} \\ \bottomrule
  \end{tabular}
  }
  \caption{Comparison in syntactic chunking. Cells show F$_1$ and standard deviations on the CoNLL-2000 test set.}
  \label{tab:chunking-result}
\end{table}

\paragraph{Generalizability}
\textit{Are our findings in NER generalizable to other tasks?}
To investigate it, we perform an additional experiment on the CoNLL-2000 dataset~\cite{tjong-kim-sang-buchholz-2000-introduction} for syntactic chunking.\footnote{The models are trained in the same way as in nested NER.}
While this task is similar to NER in terms of short-span classification, the class labels are based on syntax, not (entity) semantics.
In Table~\ref{tab:chunking-result}, the instance-based span model achieved competitive F$_1$ scores with the classifier-based one, which is consistent with the NER results.
This suggests that our findings in NER are likely to generalizable to other short-span classification tasks.

\paragraph{Future work}
One interesting line of future work is an extension of our method to span-to-span relation classification, such as SRL and coreference resolution.
Another potential direction is to apply and evaluate learned span features to downstream tasks requiring entity knowledge, such as entity linking and question answering.

\section{Conclusion}
\label{sec:conc}
We presented and investigated an instance-based learning method that learns similarity between spans.
Through NER experiments, we demonstrated that the models build by our method have (i) competitive performance with a classifier-based span model and (ii) interpretable inference process where it is easy to understand how much each training instance contributes to the predictions.

\section*{Acknowledgments}
This work was partially supported by JSPS KAKENHI Grant Number JP19H04162 and JP19K20351.
We would like to thank the members of Tohoku NLP Laboratory and the anonymous reviewers for their insightful comments.

\bibliography{acl2020}
\bibliographystyle{acl_natbib}

\newpage
\appendix
\section{Appendices}
\subsection{Experimental setup}
\label{sec:model-config}

\begin{table}[h]
  \begin{center}
  {\small
  \begin{tabular}{lr} \toprule
    Name & Value \\ \midrule
    CNN window size & 3\\
    CNN filters & 30 \\
    BiLSTM layers & 2 \\
    BiLSTM hidden units  & 100 dimensions \\
    Mini-batch size & 8 \\
    Optimization & Adam \\
    Learning rate & 0.001 \\
    Dropout ratio & \{0.1, 0.3, 0.5\} \\ \bottomrule
  \end{tabular}
  }
  \end{center}
  \caption{Hyperparameters used in the experiments.}
\label{tab:hyperparam}
\end{table}

\paragraph{Network setup}
Basically, we follow the encoder architecture proposed by~\citet{ma-hovy-2016-end}.
First, the token-encoding layer encodes each token of the input sentence $w_t \in (w_1, w_2, \dots, w^{}_T)$ to a sequence of the vector representations ${\bf w}_{1:T} = (\textbf{w}_1, \textbf{w}_2, \dots, \textbf{w}_T)$.
For the models using GloVe, we use the GloVe 100-dimensional embeddings\footnote{\url{https://nlp.stanford.edu/projects/glove/}} \cite{pennington-etal-2014-glove} and character-level CNN.
For the models using BERT,  we use the BERT-Base, Cased\footnote{\url{https://github.com/google-research/bert}} \cite{devlin-etal-2019-bert}, where we use the first subword embeddings within each token in the last layer of BERT.
During training, we fix the word embeddings (except the CNN).
Then, the encoded token representations ${\bf w}_{1:T} = (\textbf{w}_1, \textbf{w}_2, \dots, \textbf{w}_T)$ are fed to bidirectional LSTM (BiLSTM)~\cite{graves:13} for computing contextual ones $\overrightarrow{\bf h}_{1:T}$ and $\overleftarrow{\bf h}_{1:T}$.
We use $2$ layers of the stacked BiLSTMs (2 forward and 2 backward LSTMs) with 100-dimensional hidden units.
From $\overrightarrow{\bf h}_{1:T}$ and $\overleftarrow{\bf h}_{1:T}$, we create ${\bf h}^\text{lstm}_s$ for each span $s = (a, b)$ based on LSTM-minus \cite{wang-chang-2016-graph}.
For flat NER, we use the representation $\mathbf{h}^\mathrm{lstm}_s =  [\overrightarrow{\mathbf{h}}_b- \overrightarrow{\mathbf{h}}_{a-1}, \overleftarrow{\mathbf{h}}_a - \overleftarrow{\mathbf{h}}_{b+1}]$.
For nested NER, we use $\mathbf{h}^\mathrm{lstm}_s =  [\overrightarrow{\mathbf{h}}_b- \overrightarrow{\mathbf{h}}_{a-1}, \overleftarrow{\mathbf{h}}_a - \overleftarrow{\mathbf{h}}_{b+1}, \overrightarrow{\mathbf{h}}_a+ \overrightarrow{\mathbf{h}}_b, \overleftarrow{\mathbf{h}}_a+ \overleftarrow{\mathbf{h}}_b]$.
We then multiply $\mathbf{h}^\mathrm{lstm}_s$ with a weight matrix $\mathbf{W}$ and obtain the span representation: $\mathbf{h}_s = \mathbf{W} \: \mathbf{h}^\mathrm{lstm}_s$.
Finally, we use the span representation $\mathbf{h}_s$ for computing the label distribution in each model.
For efficient computation, following \citet{sohrab-miwa-2018-deep}, we enumerate all possible spans in a sentence with the sizes less than or equal to the maximum span size $L$, i.e., each span $s = (a,b)$ is satisfied with the condition $b - a < L$.
We set $L$ as $6$.

\paragraph{Hyperparameters}
Table~\ref{tab:hyperparam} lists the hyperparameters used in the experiments.
We initialize all the parameter matrices in BiLSTMs with random orthonormal matrices~\cite{saxe:13}.
Other parameters are initialized following~\newcite{glorot:10}.
We apply dropout~\cite{srivastava:14} to the token-encoding layer and the input vectors of each LSTM with dropout ratio of $\{0.1, 0.3, 0.5\}$.

\paragraph{Optimization}
To optimize the parameters, we use Adam \cite{kingma:14} with $\beta_1 = 0.9$ and $\beta_2 = 0.999$.
The initial learning rate is set to $\eta_0 = 0.001$.
The learning rate is updated on each epoch as $\eta_t = \eta_0 / (1+\rho t)$, where the decay rate is $\rho = 0.05$ and $t$ is the number of epoch completed.
A gradient clipping value is set to $5.0$~\cite{pascanu2013difficulty}.
Parameter updates are performed in mini-batches of 8.
The number of training epochs is set to 100.
We save the parameters that achieve the best F1 score on each development set and evaluated them on each test set.
Training the models takes less than one day on a single GPU, NVIDIA DGX-1 with Tesla V100.

\subsection{Feature space visualization}
\begin{figure}[h]
  \centering
  \begin{minipage}{0.49\hsize}
    \centering
    \includegraphics[height=4cm]{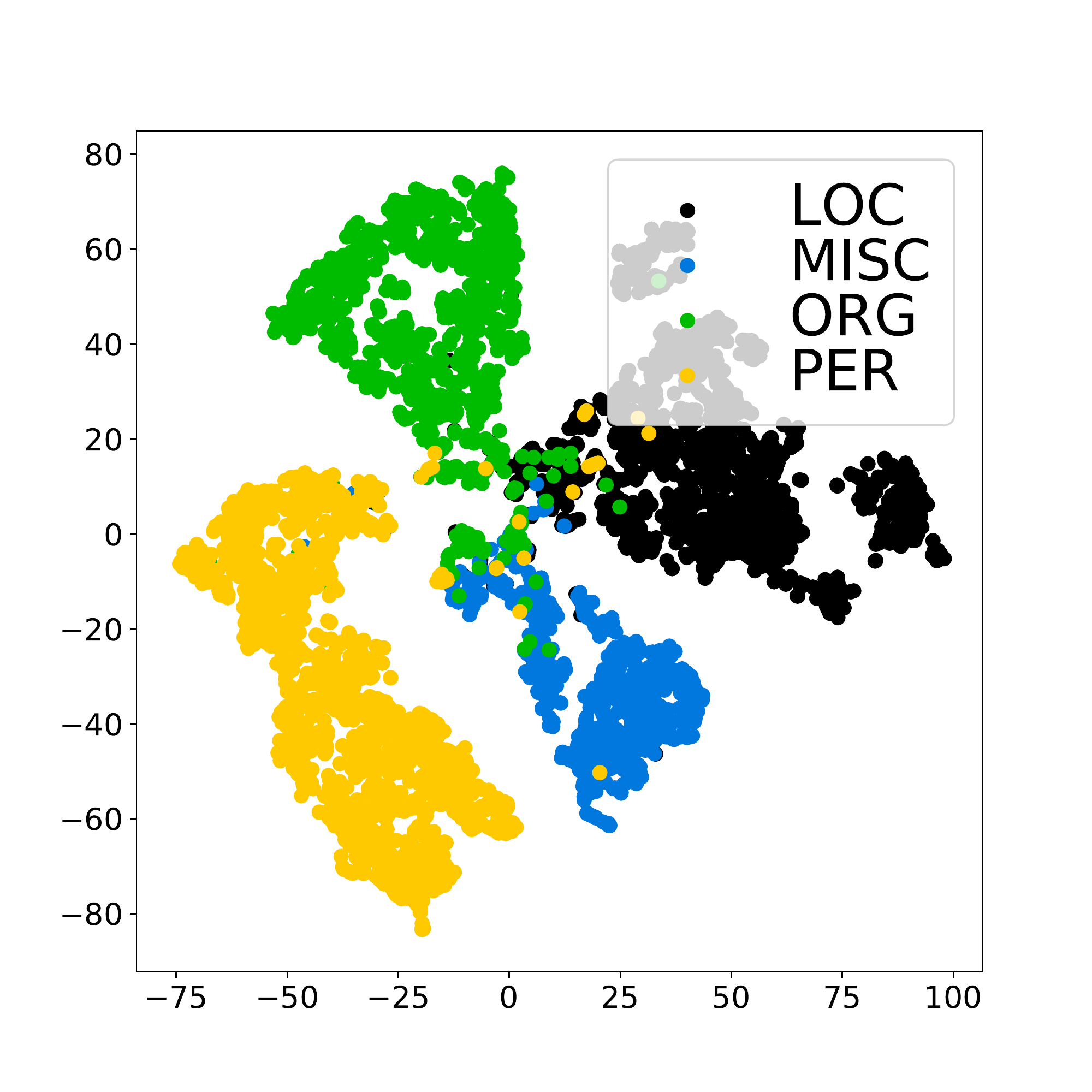}
    \vspace{-0.8cm}
    \subcaption{Classifier-based}
    \label{fig:left}
  \end{minipage}
  \begin{minipage}{0.49\hsize}
    \centering
    \includegraphics[height=4cm]{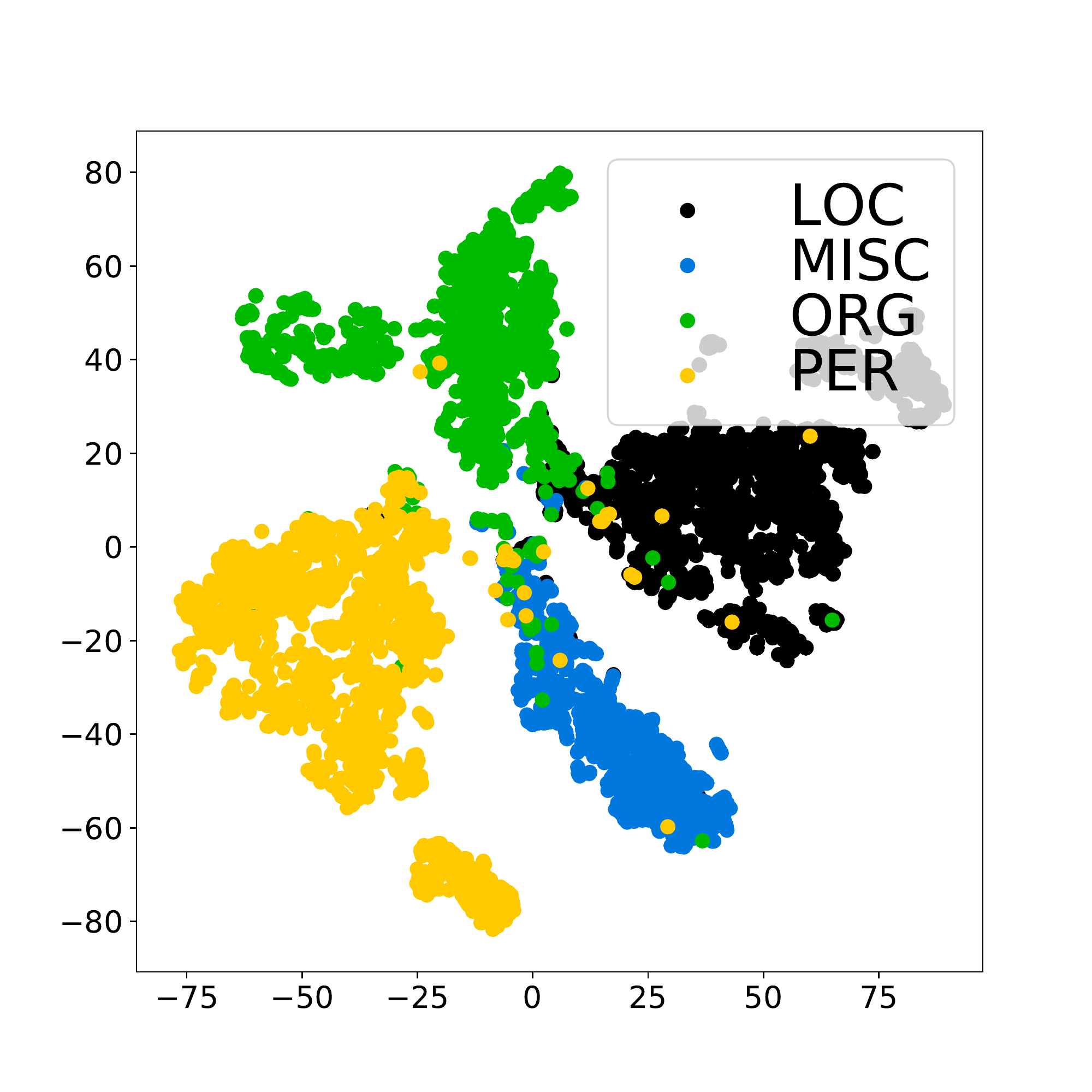}
    \vspace{-0.8cm}
    \subcaption{Instance-based}
    \label{fig:right}
  \end{minipage}
\caption{Visualization of entity span features computed by classifier-based and instance-based models.}
\label{fig:label-emb}
\end{figure}

To better understand span representations learned by our method, we observe the feature space.
Specifically, we visualize the span representations $\textbf{h}_s$ on the CoNLL-2003 development set.
Figure~\ref{fig:label-emb} visualizes two-dimensional entity span representations by t-distributed Stochastic Neighbor Embedding (t-SNE)~\cite{maaten2008visualizing}.
Both models successfully learned feature spaces where the instances with the same label come close each other.

\end{document}